\documentclass{article}

\PassOptionsToPackage{dvipsnames}{xcolor} %

\usepackage[final]{neurips_2022}

\usepackage[utf8]{inputenc} %
\usepackage[T1]{fontenc}    %
\usepackage{hyperref}       %
\usepackage{url}            %
\usepackage{booktabs}       %
\usepackage{amsfonts}       %
\usepackage{nicefrac}       %
\usepackage{microtype}      %
\usepackage{xcolor}         %

\usepackage{xspace}
\usepackage{tabularx}
\usepackage[inline]{enumitem}
\usepackage{mleftright}
\usepackage[group-separator={,}]{siunitx}
\usepackage[framemethod=tikz]{mdframed}
\usepackage{amsmath}
\usepackage{soul}
\usepackage{colortbl}
\usepackage[normalem]{ulem}
\usepackage{xr}
\usepackage{changepage}
\usepackage{placeins}
\usepackage{xurl}
\usepackage{nowidow}
\usepackage{amsmath,bm}
\usepackage{authblk}

\usepackage[setmargin=true,marginparwidth=0.8in,hide=true]{marginalia}
\usepackage{cleveref}

\title{The Effect of Data Dimensionality on Neural Network Prunability}

\makeatletter
    \renewcommand\AB@affilsepx{ \hphantom{---} \protect\Affilfont}
\makeatother

\author[1,2]{
  Zachary Ankner\thanks{Correspondence to \texttt{ankner@mit.edu}}\hphantom{x}}
\author[1]{Alex Renda}
\author[3]{Gintare Karolina Dziugaite}
\author[2]{Jonathan Frankle}
\author[1]{Tian Jin}

\affil[1]{Massachusetts Institute of Technology}
\affil[2]{MosaicML}
\affil[3]{Google Research, Brain Team}

\DeclareMathOperator{\sign}{sign}

\begin{document}

\maketitle

\begin{abstract}
     Practitioners prune neural networks for efficiency gains and generalization improvements, but few scrutinize the factors determining the \emph{prunability} of a neural network -- the maximum fraction of weights that pruning can remove without compromising the model's test accuracy. 
     In this work, we study the properties of input data that may contribute to the prunability of a neural network.

     For high dimensional input data such as images, text, and audio, the \emph{manifold hypothesis} suggests that these high dimensional inputs approximately lie on or near a significantly lower dimensional manifold.
     Prior work demonstrates that the underlying low dimensional structure of the input data may affect the sample efficiency of learning.
     In this paper, we investigate whether the low dimensional structure of the input data affects the prunability of a neural network.
\end{abstract}
\section{Introduction}

The \emph{manifold hypothesis} states that the input data for tasks involving images, texts and sounds approximately lie in a low-dimensional manifold embedded in a high dimensional space \citep{fefferman2016testing}.
Ample \NA{evidence} 
\citep{Narayanan2009OnTS, pope2021the} argue that the manifold hypothesis is connected to the sample efficiency of deep neural network models.
We investigate whether the low-dimensional manifold structure of input data may also play a role in the surprising empirical success of weight pruning -- pruning techniques can remove more than 90\% of weights in a neural network model without compromising its accuracy \citep{lecun_optimal_1990, frankle_lottery_2018, han_learning_2015}.

In this work, we empirically examine the impact of three different measures of input data dimensionality on the \emph{prunability} of a neural network model, defined as the maximum fraction of weights that a pruning technique can remove without reducing model accuracy.
We refer to the dimensionality of the model inputs as the \emph{extrinsic dimensionality}, and the dimensionality of the lower dimensional manifold that the data lies on as its \emph{intrinsic dimensionality}.
We also define a new dimensionality, \emph{task dimensionality}, which is the minimal number of intrinsic dimensions that the output of the function being learned depends on.
We develop and test three hypotheses: that the extrinsic, intrinsic, and task dimensionality each 
contribute to the prunability of neural network models 
To test these hypotheses we design three experiments: for each type of dimensionality (extrinsic, intrinsic, and task) we vary that dimensionality while holding all others fixed and observe its effect on the prunability of the neural network model.

\pagebreak
\paragraph{Contributions.}

We present the following contributions:
\begin{itemize}
     \item We find that extrinsic dimensionality \textbf{has limited effect on} neural network prunability. As extrinsic dimension increases prunability slightly increases with weak correlation.
     \item We find that intrinsic dimensionality \textbf{does affect} neural network prunability. As intrinsic dimension increases prunability decreases.
     \item We find that task dimensionality \textbf{does not affect} neural network prunability.
\end{itemize}

\section{Preliminaries}
\label{prelims}
\newcommand{\defn}[1]{\emph{\color{Black} #1}}

\paragraph{Dimensionality of data.}
Consider a classification task on $D$-dimensional inputs.
We refer to the input dimensionality $D$ as the \defn{extrinsic dimensionality}.
The \emph{manifold hypothesis} states that natural data lies on or near a low-dimensional manifold embedded in high dimensional space \citep{fefferman2016testing}.
We refer to dimensionality of the manifold that the data lies on as the \defn{intrinsic dimensionality} $d$.
Thus, according to the manifold hypothesis, $d \ll D$.
We refer to the low dimensional manifold that the inputs lie on as the \emph{intrinsic space}.

\paragraph{Pruning.}

We prune a neural network using \defn{iterative magnitude pruning (IMP)} with the weight rewinding algorithm \citep{frankle_linear_2019}, which is considered to be state-of-the-art \citep{pmlr-v139-rosenfeld21a, DBLP:journals/corr/abs-1902-09574}.
\NA{IMP begins} 

A \defn{matching subnetwork} is a sparse subnetwork with error that is equal to or lower than the error of the original dense network that the sparse subnetwork was obtained from.
The highest sparsity at which a pruned subnetwork is still matching describes the \defn{prunability} of a network with respect to a particular input data distribution and task.

\paragraph{Correlation testing.}
\label{prelim-corr}
As we are interested in determining whether varying data dimensionality affects prunability, we quantify the correlation between the ordering of data dimensionalities and prunability using the Spearman rank correlation test.
To reflect the underlying uncertainty of the observed pruning rates in the correlation coefficient, we employ a Monte Carlo method to estimate the distribution of the correlation coefficient~\citep{curran2014-spearman-estimate}.
To summarize the correlation of a set of experiments, we report both the mean and standard deviation of the correlation coefficient.
More details on how we estimate the correlation coefficient distribution are described in Appendix \ref{corr-methodology}.

\section{Extrinsic Dimensionality Has Limited Effect on Prunability}
\label{sec:extrinsic-dim}

For an image-classification model, the extrinsic dimensionality of its input data is equal to the number of pixels per image.
\NA{The difference between the extrinsic dimensionality and the intrinsic dimensionality of the input data is therefore a natural source of redundancy in the representation of the input data, which may contribute to the prunability of neural network models.}
In this section, we examine the impact of the extrinsic dimensionality of the input data on the prunability of a neural network model, while keeping other data dimensionalities fixed.

\paragraph{Methodology.}
Following the precedent of \citet{pope2021the}, we vary the extrinsic dimensionality of the input data by resizing the input image using nearest-neighbor interpolation.
For each target extrinsic dimensionality, we resize the spatial dimensions of the input images in a given dataset to the specified extrinsic dimensionality, creating a distinct dataset for each extrinsic dimensionality we study.
Subsequently, we train and prune a specified model architecture using the aforementioned dataset.
We examine the relationship between the extrinsic dimensionality of the input data and the prunability of the neural network model.
We experiment on the CIFAR-10 dataset which originally has an extrinsic dimensionality of 32 (height) x 32 (width) x 3 (channel) = 3072.
We vary its extrinsic dimensionality by resizing its spatial dimension to create four separate datasets of spatial dimension $16 \times 16, 32 \times 32, 64 \times 64,$ and $128 \times 128$.
We experiment with 3 variants of the ResNet20 model, evaluating across model widths of $16, 32,$ and $64$.
The number of weights in a convolutional neural network is independent of the spatial dimension of its input.
Therefore, in our experiments, varying the extrinsic dimensionality of the input data does not alter the number of weights present in the neural network. 

\paragraph{Results.}

\begin{figure}
     \centering
     \includegraphics[width=\textwidth]{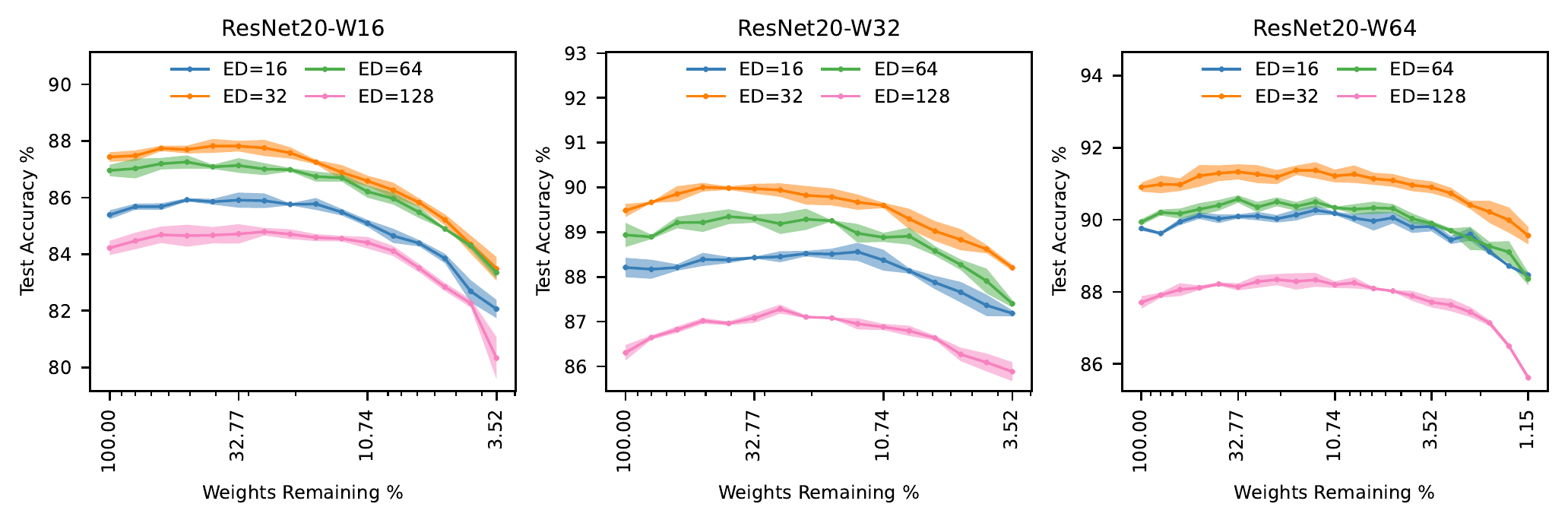}
     \caption{Test accuracy as a function of the percentage of model weights remaining for each model width and extrinsic dimensionality. \NA{Pruning across varying extrinsic dimensionalities often follows a line with similar curvature.} Legend indicates extrinsic dimension.}
     \label{fig:extrinsic}
\end{figure}

In Figure~\ref{fig:extrinsic} we plot the model's test accuracy as a function of the percentage of weights remaining for each of the combinations of model width and extrinsic dimensionality.
We find that, while the extrinsic dimensionality has an impact on the initial accuracy of the dense model
We also find that there is no clear relationship between the ordering of extrinsic dimensionalities and the ordering of their corresponding pruning rates.
See Table~\ref{tab:extrinsic} for numerical results.
To quantify the weak correlation between the extrinsic dimensionality and prunability, we compute the Spearman rank correlation between the extrinsic dimensionality and the percentage of weights remaining in the smallest matching subnetworks.
Across the different model widths, we find an average rank correlation coefficient of -0.22$\pm{0.03}$.
This suggests that while prunability increases as the extrinsic dimensionality increases, it is only a weakly correlated relationship.

\paragraph{Takeaway.}
We find that increasing the extrinsic dimensionality of input data by upsampling their spatial dimensions is weakly correlated with increased prunability of neural network models.

\section{Intrinsic Dimensionality Does Affect Prunability}
\label{sec:intrinsic}
In this section, we examine the effect of the intrinsic dimension on neural network prunability.
The intrinsic dimensionality of the data is the dimensionality of the low dimensional manifold the data lies on.
Thus, we hypothesize that as the intrinsic dimensionality of the data increases the prunability of the neural network will decrease as there are fewer redundant dimensions in the data.

\paragraph{Methodology.}
We follow the precedent of \cite{pope2021the} to generate datasets of varying intrinsic dimensionalities.
Concretely, we use a class conditional Generative Adversarial Network (GAN), specifically BigGAN~\citep{brock2018-biggan}, to generate images as inputs which are labeled by the class they are conditioned on.
The GAN network maps vectors in a latent space to natural images.
The dimensionality of the vectors sampled within the latent space is therefore an upper bound on the intrinsic dimensionality of the generated dataset.
To vary the intrinsic dimensionality, we mask out the corresponding number of latent dimensions from each vector sampled in the latent space by setting them to 0, before feeding them to the GAN network.
For this experiment, we fix the extrinsic dimensionality of the images to be $32 \times 32 \times 3$ and vary the intrinsic dimensionality in $\{16, 32, 64, 128\}$.
We experiment with 3 variants of the ResNet20 model, evaluating across model widths of $8, 16,$ and $32$.

\paragraph{Results.}

\begin{figure}
     \centering
     \includegraphics[width=\textwidth]{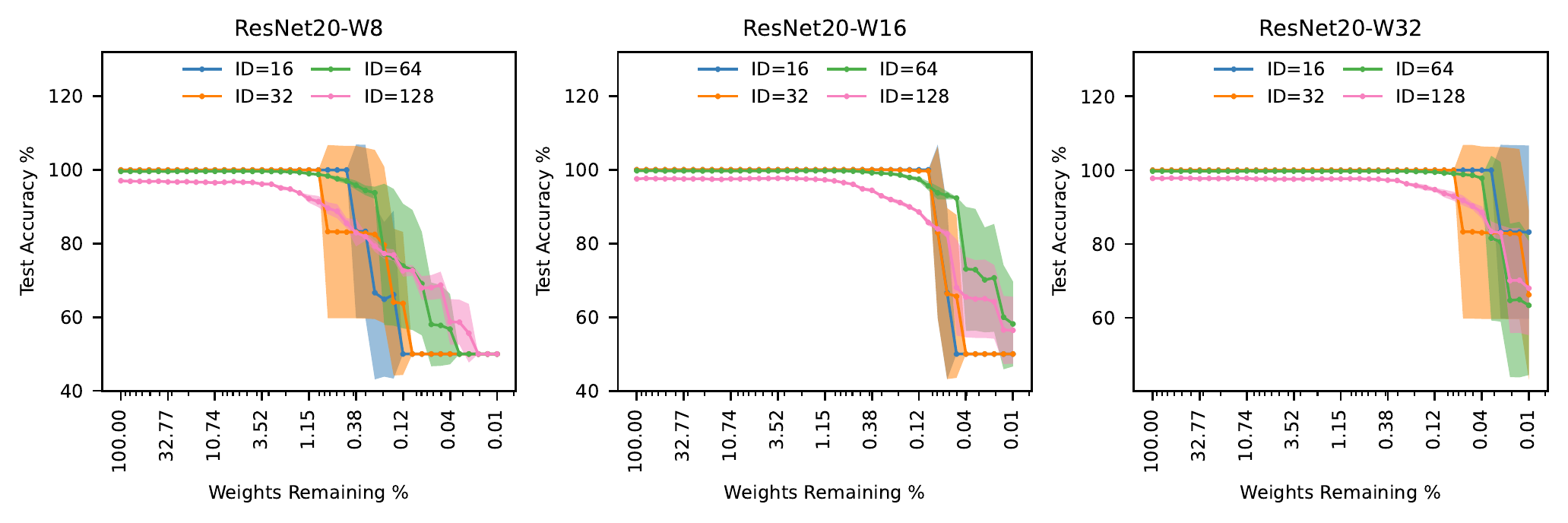}
     \caption{Test accuracy as a function of the percentage of model weights remaining for each model width and intrinsic dimension. Lower intrinsic dimension corresponds to pruned model performance matching dense accuracy for greater amounts of pruning. Legend indicates intrinsic dimension.}
     \label{fig:intrinsic}
\end{figure}

\pagebreak
In Figure~\ref{fig:intrinsic} we plot the pruned model's test accuracy as a function of the percentage of weights remaining for each of the combinations of model width and intrinsic dimensionality.
We find that models pruned on datasets with lower intrinsic dimensionalities can be pruned to fewer weights remaining without compromising accuracy compared with the same models pruned on datasets with higher intrinsic dimensionalities.
See Table~\ref{tab:intrinsic} for numerical results.
To quantify the correlation between the intrinsic dimensionality and prunability, we compute the Spearman rank correlation between the intrinsic dimensionality and the percentage of weights remaining in the smallest matching subnetworks.
Averaged across model widths, we find a high rank correlation coefficient of 0.74$\pm{0.01}$.
This correlation supports our observation that as intrinsic dimensionality increases, the percentage of weights remaining in the smallest matching subnetworks also increases.

\paragraph{Takeaway.}
For the classification datasets we generate using a GAN model, we find that the intrinsic dimensionality of the input data affects prunability.
The lower the intrinsic dimensionality of input data, the larger the fraction of model weights that can be pruned without compromising test accuracy.

\section{Task Dimensionality Does Not Affect Prunability}

For image classification tasks, the label of an image often depends only on a subset of the features present in the image.
For example, the background of an image is often irrelevant to the classification task.
Does the prunability of a neural network model depend on the fraction of the input features that contribute to the task output?
In this section, we formalize the intuition behind the number of input features that contribute to the task output and examine its influence on prunability.

\paragraph{Methodology.}
For this experiment, we assume that the label assigned to a datapoint is a function of its representation in the intrinsic space.
Then, for a labeling function $f$, we define the \emph{task dimensionality} as the number of intrinsic dimensions that the output of $f$ \NA{depends on} 
To investigate the impact the task dimensionality has on prunability, we create a synthetic dataset encoding a linear classification task as follows.

To construct inputs, we first randomly sample $d-$dimensional vectors $\bm z$ 
We randomly sample a $D$ by $d$ matrix and set the $D$ dimensional input vectors $\bm x$ = $A\bm z$.
Each component of $A$ is sampled i.i.d. from a uniform distribution $\mathcal{U}(-1,1)$.
By construction, such input vectors have intrinsic dimensionality of $d$, and extrinsic dimensionality of $D$.

To label inputs, we create hyperplanes defining classification boundaries in the intrinsic space.
We sample and fix a random $d-$dimensional vector $\bm h$, representing the normal vector to the plane wherein each component is sampled i.i.d. from a uniform distribution $\mathcal{U}(-1,1)$.
Then, we randomly select $t$ components of this hyperplane-defining normal vector to retain and zero out the remaining components.
We define the label of an input vector $\bm x$ as $y = \sign(\bm h^\top \bm x)$.
That is, we label each input vector based on which side of the hyperplane its corresponding representation in the intrinsic space lies on.
Zeroing out a component in $\bm h$ removes the dependence of classification output on the corresponding intrinsic dimension.
This classification task has a task dimensionality of $t$.

For our experiments, we vary the task dimensionality $t \in \{16, 32, 64, 128\}$ while we fix the intrinsic dimensionality $d=128$, and extrinsic dimensionality $D = 1024$.
We train and prune a 2-layer neural network model consisting of 2 linear layers.
We use MLP-P-Q to denote such 2-layer models, where P and Q correspond to the number of features in the first and second hidden layers respectively.
For each task dimension, we vary (P,Q) in ${(256, 128), (512, 256), (1024, 512)}$.

\begin{figure}
     \centering
     \includegraphics[width=\textwidth]{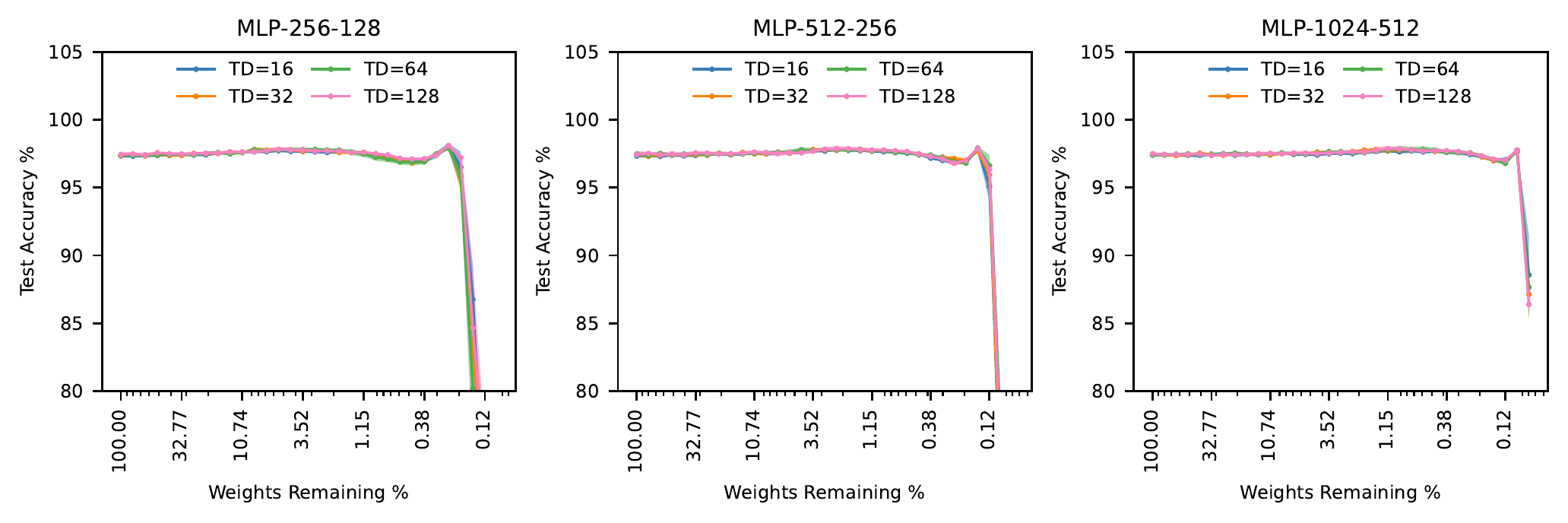}
     \caption{Test accuracy as a function of the percentage of model weights remaining for each model width and task dimension. For most runs the \NA{pruning curves} are indistinguishable and overlap highly. Legend indicates task dimension.}
     \label{fig:task}
\end{figure}

\paragraph{Results.}
In \Cref{fig:task} we plot the test accuracies of pruned models as a function of the percentage of weights that remain in the pruned models.
We find that across all model architectures we examine, the prunability of the model does not correlate with task dimensionality of the labeling function.
With one exception (MLP-256-128 with TD=16), for each dense model, the sparsities of its smallest matching subnetworks that pruning produces across all task dimensionalities are within one standard deviation of each other.
These results suggest that the task dimensionality does not affect the prunability of a neural network model.
See Table~\ref{tab:task} for numerical results.
To quantify the correlation between the task dimensionality and prunability, we compute the Spearman rank correlation between the task dimensionality and the fractions of weights that remain in the smallest matching subnetworks.
We obtain a correlation coefficient of 0.01$\pm{0.03}$ suggesting that there is no correlation between task dimensionality and prunability.

\paragraph{Takeaway.}
For the synthetic linear classification tasks we construct, we find that task dimensionality does not correlate with the prunability of a neural network model.

\section{Related Work}
\paragraph{Pruning.}
Practitioners use pruning for many reasons: pruning may improve model generalization and explainability (e.g., \citep{lecun_optimal_1990, hassibi_optimal_1993}); pruning also may reduce memory footprint and improve computational efficiency of model training or inference (e.g., \citep{han_learning_2015, frankle_lottery_2018, renda_comparing_2020, MLSYS2020_d2ddea18,Lebedev_2016_CVPR,molchanov2017pruning,NIPS2017_c5dc3e08,DBLP:journals/corr/abs-1711-05908,baykal2018datadependent,lee2018snip,wang2019eigen,DBLP:journals/corr/abs-2001-00218,Lee2020A}).
While a variety of pruning techniques exist, we choose to study iterative magnitude pruning as it achieves state-of-the-art trade-off between model size and accuracy \citep{pmlr-v139-rosenfeld21a, DBLP:journals/corr/abs-1902-09574}.

\paragraph{Cause of prunability.}
Despite the widespread empirical success of pruning for deep neural network models, there has been little analysis on the cause of prunability -- why and when can pruning remove substantial fraction of weights in a given model without compromising its accuracy?
\citet{lecun_optimal_1990, hassibi_optimal_1993, han_learning_2015} attributed the  prunability of neural network models to bias-variance trade-off: as increasing number of weights are pruned, model variance decreases while model bias increases.
A smaller model size may therefore exist with the same overall test accuracy but different decomposition of error into bias and variance.
\citet{frankle_lottery_2018} showed the existence of sparse and trainable subnetworks at initialization, thereby conjecturing that SGD seeks out and trains a subset of well-initialized weights.
Outside this select subset, weights are therefore redundant.
\citet{pmlr-v80-arora18b} showed that in deep neural network models, each layer's output is robust against noise injected at preceding layers.
They showed that this form of noise stability implies prunability of neural network model.

\section{Discussion}

In this section we further discuss our results and limitations.

\paragraph{Falsified Hypotheses.}
We begin our work with a motivating conjecture that the true low dimensional structure of input data contributes to the surprising empirical success of neural network pruning.
Based on this conjecture, we develop 3 hypotheses: that extrinsic, intrinsic, and task dimensionality affect prunability.
Our empirical observations falsify the hypothesis that task dimensionality affects prunability, and shows that the correlation between extrinsic dimensionality and prunability is weak.
Together we present a set of positive and negative results that advances our understanding of the empirical success of neural network pruning.

\paragraph{Experimental Methodology.}
In this work, we examine the empirical success of  neural network pruning from a novel perspective -- we investigate how properties of the input data affect prunability.
To this end, we design and validate several experimental methodologies.
However, this work only presents a preliminary empirical analysis.
An important next step is to derive a theoretical framework that unites the observations made across the various types of data dimensionalities studied in this work.
Our empirical analysis lays the groundwork for such future studies.

\section{Conclusion}
To the best of our knowledge, we present the first study that systematically studies the effects of several properties of the input data distribution on prunability.
We study how the extrinsic, intrinsic, and task dimensionality of the input data affects neural network prunability.
Our results empirically demonstrate that while intrinsic dimensionality affects prunability, extrinsic and task dimensionality have little to no effect on prunability.

\nocite{frankle_lottery_2018}

\bibliographystyle{plainnat}
\bibliography{references}

\begin{thebibliography}{27}
\providecommand{\natexlab}[1]{#1}
\providecommand{\url}[1]{\texttt{#1}}
\expandafter\ifx\csname urlstyle\endcsname\relax
  \providecommand{\doi}[1]{doi: #1}\else
  \providecommand{\doi}{doi: \begingroup \urlstyle{rm}\Url}\fi

\bibitem[Arora et~al.(2018)Arora, Ge, Neyshabur, and Zhang]{pmlr-v80-arora18b}
Sanjeev Arora, Rong Ge, Behnam Neyshabur, and Yi~Zhang.
\newblock Stronger generalization bounds for deep nets via a compression
  approach.
\newblock In Jennifer Dy and Andreas Krause, editors, \emph{Proceedings of the
  35th International Conference on Machine Learning}, volume~80 of
  \emph{Proceedings of Machine Learning Research}, pages 254--263. PMLR, 10--15
  Jul 2018.
\newblock URL \url{https://proceedings.mlr.press/v80/arora18b.html}.

\bibitem[Baykal et~al.(2019)Baykal, Liebenwein, Gilitschenski, Feldman, and
  Rus]{baykal2018datadependent}
Cenk Baykal, Lucas Liebenwein, Igor Gilitschenski, Dan Feldman, and Daniela
  Rus.
\newblock Data-dependent coresets for compressing neural networks with
  applications to generalization bounds.
\newblock In \emph{International Conference on Learning Representations}, 2019.
\newblock URL \url{https://openreview.net/forum?id=HJfwJ2A5KX}.

\bibitem[Blalock et~al.(2020)Blalock, Gonzalez~Ortiz, Frankle, and
  Guttag]{MLSYS2020_d2ddea18}
Davis Blalock, Jose~Javier Gonzalez~Ortiz, Jonathan Frankle, and John Guttag.
\newblock What is the state of neural network pruning?
\newblock In I.~Dhillon, D.~Papailiopoulos, and V.~Sze, editors,
  \emph{Proceedings of Machine Learning and Systems}, volume~2, pages 129--146,
  2020.
\newblock URL
  \url{https://proceedings.mlsys.org/paper/2020/file/d2ddea18f00665ce8623e36bd4e3c7c5-Paper.pdf}.

\bibitem[Brock et~al.(2018)Brock, Donahue, and Simonyan]{brock2018-biggan}
Andrew Brock, Jeff Donahue, and Karen Simonyan.
\newblock Large scale gan training for high fidelity natural image synthesis.
\newblock \emph{arXiv preprint arXiv:1809.11096}, 2018.

\bibitem[Cayton(2005)]{Cayton2005AlgorithmsFM}
Lawrence Cayton.
\newblock Algorithms for manifold learning.
\newblock 2005.

\bibitem[Curran(2014)]{curran2014-spearman-estimate}
Peter~A Curran.
\newblock Monte carlo error analyses of spearman's rank test.
\newblock \emph{arXiv preprint arXiv:1411.3816}, 2014.

\bibitem[Dong et~al.(2017)Dong, Chen, and Pan]{NIPS2017_c5dc3e08}
Xin Dong, Shangyu Chen, and Sinno Pan.
\newblock Learning to prune deep neural networks via layer-wise optimal brain
  surgeon.
\newblock In I.~Guyon, U.~Von Luxburg, S.~Bengio, H.~Wallach, R.~Fergus,
  S.~Vishwanathan, and R.~Garnett, editors, \emph{Advances in Neural
  Information Processing Systems}, volume~30. Curran Associates, Inc., 2017.
\newblock URL
  \url{https://proceedings.neurips.cc/paper/2017/file/c5dc3e08849bec07e33ca353de62ea04-Paper.pdf}.

\bibitem[Fefferman et~al.(2016)Fefferman, Mitter, and
  Narayanan]{fefferman2016testing}
Charles Fefferman, Sanjoy Mitter, and Hariharan Narayanan.
\newblock Testing the manifold hypothesis.
\newblock \emph{Journal of the American Mathematical Society}, 29\penalty0
  (4):\penalty0 983--1049, 2016.

\bibitem[Frankle and Carbin(2019)]{frankle_lottery_2018}
Jonathan Frankle and Michael Carbin.
\newblock The lottery ticket hypothesis: Finding sparse, trainable neural
  networks.
\newblock In \emph{International Conference on Learning Representations}, 2019.
\newblock URL \url{https://openreview.net/forum?id=rJl-b3RcF7}.

\bibitem[Frankle et~al.(2020)Frankle, Dziugaite, Roy, and
  Carbin]{frankle_linear_2019}
Jonathan Frankle, Gintare~Karolina Dziugaite, Daniel~M. Roy, and Michael
  Carbin.
\newblock Linear mode connectivity and the lottery ticket hypothesis.
\newblock In \emph{Proceedings of the 37th International Conference on Machine
  Learning}, ICML'20. JMLR.org, 2020.

\bibitem[Gale et~al.(2019)Gale, Elsen, and
  Hooker]{DBLP:journals/corr/abs-1902-09574}
Trevor Gale, Erich Elsen, and Sara Hooker.
\newblock The state of sparsity in deep neural networks.
\newblock \emph{CoRR}, abs/1902.09574, 2019.
\newblock URL \url{http://arxiv.org/abs/1902.09574}.

\bibitem[Ham et~al.(2004)Ham, Lee, Mika, and
  Sch\"{o}lkopf]{10.1145/1015330.1015417}
Jihun Ham, Daniel~D. Lee, Sebastian Mika, and Bernhard Sch\"{o}lkopf.
\newblock A kernel view of the dimensionality reduction of manifolds.
\newblock In \emph{Proceedings of the Twenty-First International Conference on
  Machine Learning}, ICML '04, page~47, New York, NY, USA, 2004. Association
  for Computing Machinery.
\newblock ISBN 1581138385.
\newblock \doi{10.1145/1015330.1015417}.
\newblock URL \url{https://doi.org/10.1145/1015330.1015417}.

\bibitem[Han et~al.(2015)Han, Pool, Tran, and Dally]{han_learning_2015}
Song Han, Jeff Pool, John Tran, and William Dally.
\newblock Learning both weights and connections for efficient neural network.
\newblock In C.~Cortes, N.~Lawrence, D.~Lee, M.~Sugiyama, and R.~Garnett,
  editors, \emph{Advances in Neural Information Processing Systems}, volume~28.
  Curran Associates, Inc., 2015.
\newblock URL
  \url{https://proceedings.neurips.cc/paper/2015/file/ae0eb3eed39d2bcef4622b2499a05fe6-Paper.pdf}.

\bibitem[Hassibi et~al.(1993)Hassibi, Stork, and Wolff]{hassibi_optimal_1993}
B.~Hassibi, D.G. Stork, and G.J. Wolff.
\newblock Optimal brain surgeon and general network pruning.
\newblock In \emph{{IEEE} International Conference on Neural Networks}, pages
  293--299 vol.1, 1993.
\newblock \doi{10.1109/ICNN.1993.298572}.

\bibitem[Lebedev and Lempitsky(2016)]{Lebedev_2016_CVPR}
Vadim Lebedev and Victor Lempitsky.
\newblock Fast convnets using group-wise brain damage.
\newblock In \emph{Proceedings of the IEEE Conference on Computer Vision and
  Pattern Recognition (CVPR)}, June 2016.

\bibitem[{LeCun} et~al.(1990){LeCun}, Denker, and Solla]{lecun_optimal_1990}
Yann {LeCun}, John Denker, and Sara Solla.
\newblock Optimal brain damage.
\newblock In D.~Touretzky, editor, \emph{Advances in Neural Information
  Processing Systems}, volume~2. Morgan-Kaufmann, 1990.
\newblock URL
  \url{https://proceedings.neurips.cc/paper/1989/file/6c9882bbac1c7093bd25041881277658-Paper.pdf}.

\bibitem[Lee et~al.(2019)Lee, Ajanthan, and Torr]{lee2018snip}
Namhoon Lee, Thalaiyasingam Ajanthan, and Philip Torr.
\newblock {SNIP}: {SINGLE}-{SHOT} {NETWORK} {PRUNING} {BASED} {ON} {CONNECTION}
  {SENSITIVITY}.
\newblock In \emph{International Conference on Learning Representations}, 2019.
\newblock URL \url{https://openreview.net/forum?id=B1VZqjAcYX}.

\bibitem[Lee et~al.(2020)Lee, Ajanthan, Gould, and Torr]{Lee2020A}
Namhoon Lee, Thalaiyasingam Ajanthan, Stephen Gould, and Philip H.~S. Torr.
\newblock A signal propagation perspective for pruning neural networks at
  initialization.
\newblock In \emph{International Conference on Learning Representations}, 2020.
\newblock URL \url{https://openreview.net/forum?id=HJeTo2VFwH}.

\bibitem[Molchanov et~al.(2017)Molchanov, Tyree, Karras, Aila, and
  Kautz]{molchanov2017pruning}
Pavlo Molchanov, Stephen Tyree, Tero Karras, Timo Aila, and Jan Kautz.
\newblock Pruning convolutional neural networks for resource efficient
  inference.
\newblock In \emph{International Conference on Learning Representations}, 2017.
\newblock URL \url{https://openreview.net/forum?id=SJGCiw5gl}.

\bibitem[Narayanan and Mitter(2010)]{NIPS2010_8a1e808b}
Hariharan Narayanan and Sanjoy Mitter.
\newblock Sample complexity of testing the manifold hypothesis.
\newblock In J.~Lafferty, C.~Williams, J.~Shawe-Taylor, R.~Zemel, and
  A.~Culotta, editors, \emph{Advances in Neural Information Processing
  Systems}, volume~23. Curran Associates, Inc., 2010.
\newblock URL
  \url{https://proceedings.neurips.cc/paper/2010/file/8a1e808b55fde9455cb3d8857ed88389-Paper.pdf}.

\bibitem[Narayanan and Niyogi(2009)]{Narayanan2009OnTS}
Hariharan Narayanan and Partha Niyogi.
\newblock On the sample complexity of learning smooth cuts on a manifold.
\newblock In \emph{COLT}, 2009.

\bibitem[Pope et~al.(2021)Pope, Zhu, Abdelkader, Goldblum, and
  Goldstein]{pope2021the}
Phil Pope, Chen Zhu, Ahmed Abdelkader, Micah Goldblum, and Tom Goldstein.
\newblock The intrinsic dimension of images and its impact on learning.
\newblock In \emph{International Conference on Learning Representations}, 2021.
\newblock URL \url{https://openreview.net/forum?id=XJk19XzGq2J}.

\bibitem[Renda et~al.(2020)Renda, Frankle, and Carbin]{renda_comparing_2020}
Alex Renda, Jonathan Frankle, and Michael Carbin.
\newblock Comparing rewinding and fine-tuning in neural network pruning.
\newblock In \emph{International Conference on Learning Representations}, 2020.
\newblock URL \url{https://openreview.net/forum?id=S1gSj0NKvB}.

\bibitem[Rosenfeld et~al.(2021)Rosenfeld, Frankle, Carbin, and
  Shavit]{pmlr-v139-rosenfeld21a}
Jonathan~S Rosenfeld, Jonathan Frankle, Michael Carbin, and Nir Shavit.
\newblock On the predictability of pruning across scales.
\newblock In Marina Meila and Tong Zhang, editors, \emph{Proceedings of the
  38th International Conference on Machine Learning}, volume 139 of
  \emph{Proceedings of Machine Learning Research}, pages 9075--9083. PMLR,
  18--24 Jul 2021.
\newblock URL \url{https://proceedings.mlr.press/v139/rosenfeld21a.html}.

\bibitem[Serra et~al.(2020)Serra, Kumar, and
  Ramalingam]{DBLP:journals/corr/abs-2001-00218}
Thiago Serra, Abhinav Kumar, and Srikumar Ramalingam.
\newblock Lossless compression of deep neural networks.
\newblock In Emmanuel Hebrard and Nysret Musliu, editors, \emph{Integration of
  Constraint Programming, Artificial Intelligence, and Operations Research},
  pages 417--430, Cham, 2020. Springer International Publishing.
\newblock ISBN 978-3-030-58942-4.

\bibitem[Wang et~al.(2019)Wang, Grosse, Fidler, and Zhang]{wang2019eigen}
Chaoqi Wang, Roger Grosse, Sanja Fidler, and Guodong Zhang.
\newblock {E}igen{D}amage: Structured pruning in the {K}ronecker-factored
  eigenbasis.
\newblock In \emph{Proceedings of the 36th International Conference on Machine
  Learning}, volume~97, pages 6566--6575. PMLR, 2019.
\newblock URL \url{http://proceedings.mlr.press/v97/wang19g.html}.

\bibitem[Yu et~al.(2018)Yu, Li, Chen, Lai, Morariu, Han, Gao, Lin, and
  Davis]{DBLP:journals/corr/abs-1711-05908}
Ruichi Yu, Ang Li, Chun-Fu Chen, Jui-Hsin Lai, Vlad~I Morariu, Xintong Han,
  Mingfei Gao, Ching-Yung Lin, and Larry~S Davis.
\newblock Nisp: Pruning networks using neuron importance score propagation.
\newblock In \emph{Proceedings of the IEEE conference on computer vision and
  pattern recognition}, pages 9194--9203, 2018.

\end{thebibliography}

\appendix
\section{Appendix}

\subsection{Acknowledgement}
This work was supported in part by a Facebook Research Award, the MIT-IBM Watson AI-LAB, Google's Tensorflow Research Cloud, and the Office of Naval Research (ONR N00014-17-1-2699).

\subsection{Correlation of Input Dimensionality and Prunability}
In this section we describe how we compute the Spearman rank correlation coefficient for each setting, and plot the correlation between prunability and data dimensionality.

\paragraph{Methodology}
\label{corr-methodology}
As outlined in Section~\ref{prelim-corr}, we compute the Spearman rank correlation coefficient between the minimum fraction of weights remaining and the dimensionality of the data.
To capture the uncertainty in our experiments in the rank correlation coefficient, we simulate experiment outcomes using the mean and standard deviation of the fraction of weights remaining.
For each rollout, we generate 1000 potential experimental outcomes by randomly selecting a data dimensionality and then sampling an outcome from a normal distribution with the same mean and standard deviation as the observed trial.
We then compute the rank correlation coefficient based on the 1000 sampled outcomes.
To collect the distribution of correlation coefficients, we perform 5000 rollouts.

\paragraph{Results}

\begin{figure}[!hb]
     \centering
     \includegraphics[width=0.32\textwidth]{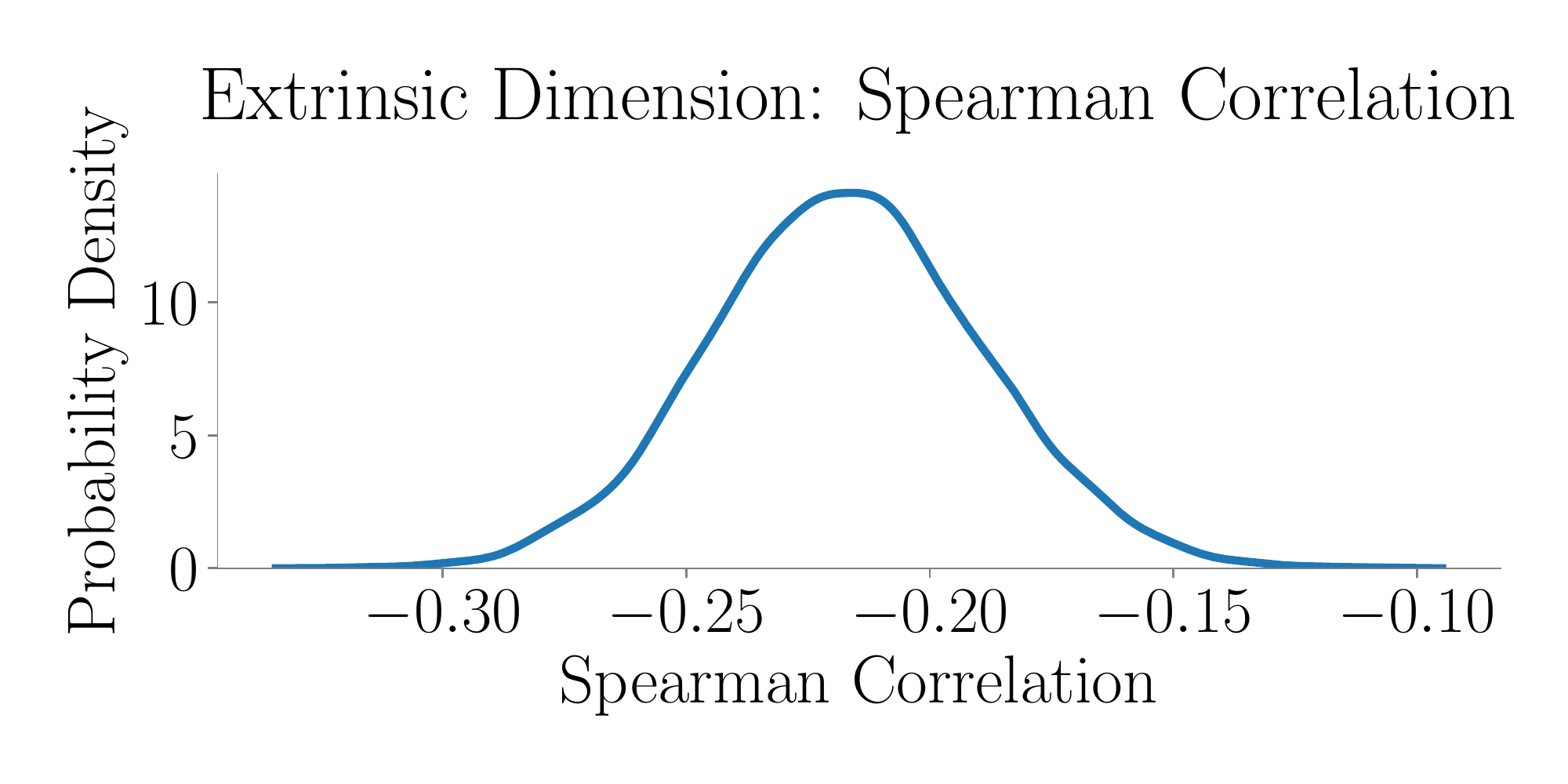}
     \includegraphics[width=0.32\textwidth]{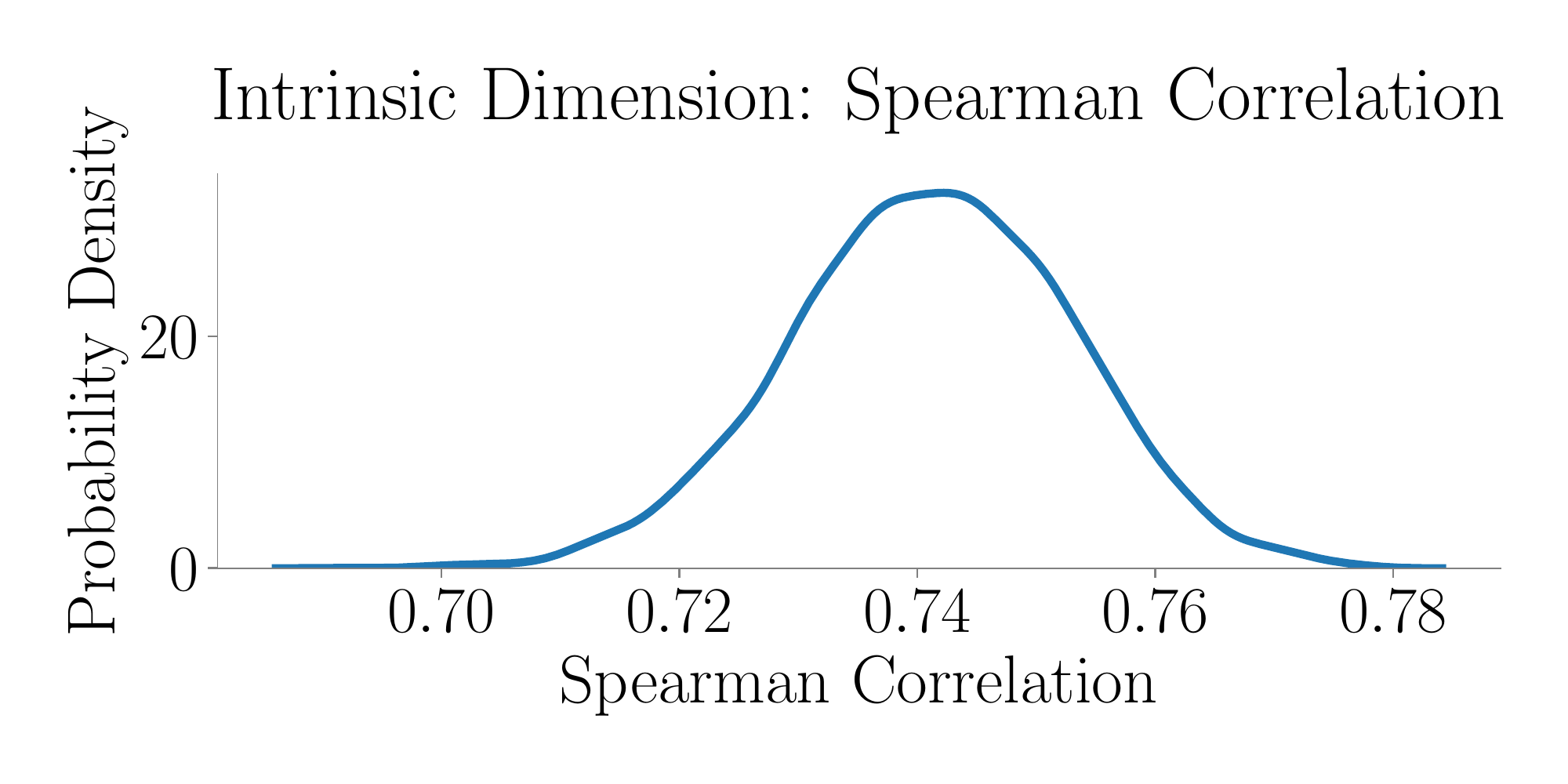}
     \includegraphics[width=0.32\textwidth]{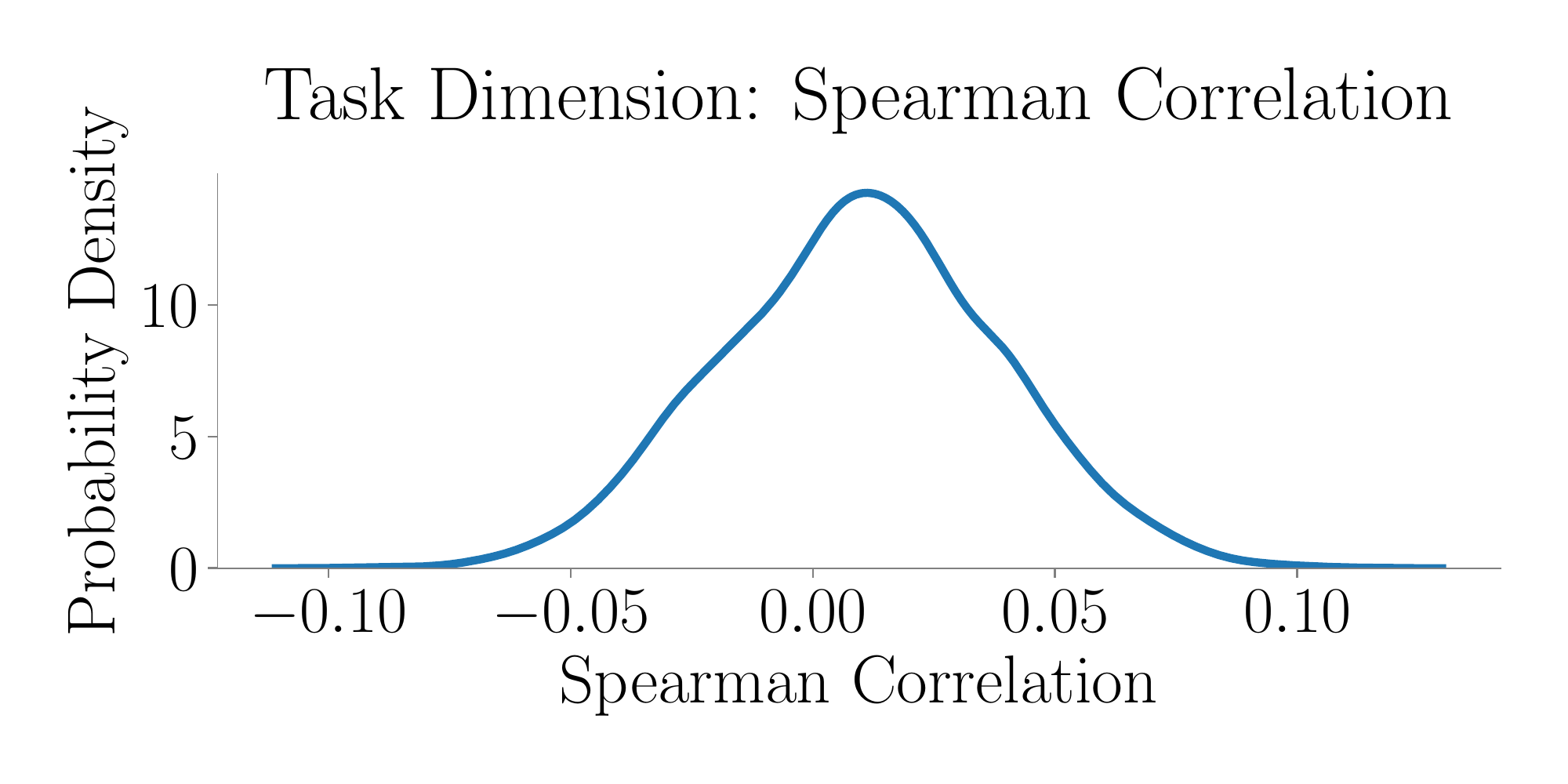}
     \caption{Distribution of Spearman rank correlation coefficients between the data dimensionality and the percentage of model weights remaining at matching sparse accuracy.}
     \label{fig:order-correlation}
\end{figure}

As can be seen in Figure~\ref{fig:order-correlation}, the intrinsic dimension experiments have a high positive correlation between input dimensionality and percentage of weights remaining, while the extrinsic and task dimension experiments have weak and no correlation respectively.
This can be interpreted as higher intrinsic dimensions are correlated with a greater percentage of weights remaining and thus lower sparsity, while the extrinsic and task dimensions do not significantly influence the prunability.

\subsection{Sparsity Rates}
In this section we report the full set of results on pruning rates across extrinsic, intrinsic, and task dimensionality (Table~\ref{tab:extrinsic},\ref{tab:intrinsic},\ref{tab:task}).

\begin{table}[htb]
     \caption{Minimum percentage of weights remaining where the sparse model matches the dense model accuracy. Results reported for each model width and extrinsic dimension.}
     \label{tab:extrinsic}
     \centering
     \begin{tabular}{lcc}
          \toprule
          Model     & Extrinsic dimension     & Weights remaining (\%) \\
          \midrule
          ResNet20-W16 & 16   & $15.86\pm{0.71}$     \\
               & 32   & $19.15\pm{1.96}$      \\
               & 64   & $18.38\pm{1.70}$  \\
               & 128  & $14.53\pm{1.30}$  \\
          \midrule
          ResNet20-W32 & 16   & $13.37\pm{1.33}$     \\
               & 32   & $14.03\pm{0.59}$      \\
               & 64   & $13.26\pm{0.50}$  \\
               & 128  & $11.76\pm{1.12}$  \\
          \midrule
          ResNet20-W64 & 16   & $10.66\pm{0.69}$     \\
               & 32   & $10.17\pm{0.00}$      \\
               & 64   & $10.17\pm{0.00}$  \\
               & 128  & $9.39\pm{0.25}$  \\
          \bottomrule
     \end{tabular}
\end{table}

\begin{table}[htb]
     \caption{Minimum percentage of weights remaining where the sparse model matches the dense model accuracy. Results reported for each model width and intrinsic dimension.}
     \label{tab:intrinsic}
     \centering
     \begin{tabular}{lcc}
       \toprule
       Model     & Intrinsic dimension     & Weights remaining (\%) \\
       \midrule
       ResNet20-W8 & 16   & $6.32\pm{0.23}$     \\
            & 32   & $7.73\pm{0.00}$      \\
            & 64   & $9.64\pm{0.86}$  \\
            & 128  & $70.85\pm{41.22}$  \\
       \midrule
       ResNet20-W16 & 16   & $5.43\pm{0.41}$     \\
            & 32   & $6.16\pm{0.11}$      \\
            & 64   & $23.34\pm{22.58}$  \\
            & 128  & $8.87\pm{0.23}$  \\
       \midrule
       ResNet20-W32 & 16   & $4.37\pm{0.10}$     \\
            & 32   & $5.18\pm{0.29}$      \\
            & 64   & $6.09\pm{0.29}$  \\
            & 128  & $9.36\pm{2.49}$  \\
       \bottomrule
     \end{tabular}
\end{table}

\begin{table}[ht!]
     \caption{Minimum percentage of weights remaining where the sparse model matches the dense model accuracy. Results reported for each model width and task dimension.}
     \label{tab:task}
     \centering
     \begin{tabular}{lcc}
       \toprule
       Model     & Task dimension     & Weights remaining (\%) \\
       \midrule
       MLP-256-128 & 16   & $5.65\pm{0.09}$     \\
            & 32   & $5.79\pm{0.00}$      \\
            & 64   & $5.79\pm{0.00}$  \\
            & 128  & $5.72\pm{0.09}$  \\
       \midrule
       MLP-512-256 & 16   & $5.40\pm{0.00}$     \\
            & 32   & $5.40\pm{0.00}$      \\
            & 64   & $5.34\pm{0.08}$  \\
            & 128  & $5.40\pm{0.00}$  \\
       \midrule
       MLP-1024-512 & 16   & $5.06\pm{0.00}$     \\
            & 32   & $5.06\pm{0.00}$      \\
            & 64   & $5.06\pm{0.00}$  \\
            & 128  & $5.06\pm{0.00}$  \\
       \bottomrule
     \end{tabular}
\end{table}

\end{document}